\documentclass[conference]{IEEEtran}

\usepackage{graphicx}
\usepackage{amsmath}
\usepackage{amssymb}
\usepackage{cite}
\usepackage{booktabs}
\usepackage{multirow}
\usepackage{array}
\usepackage{caption}
\usepackage{subcaption}
\usepackage{ragged2e}  % for \justifying
\usepackage{cuted}     % for one-column environment

\makeatletter
\renewcommand\thesection{\arabic{section}}

\def\section{\@startsection{section}{1}{\z@}%
  {2.0ex plus 1ex minus .2ex}% space ABOVE section
  {2.8ex plus .5ex}% *** MUCH larger space BELOW section ***
  {\normalfont\large\bfseries}}

\renewcommand\thesubsection{\thesection.\arabic{subsection}}

\def\@sect#1#2#3#4#5#6[#7]#8{%
  \ifnum #2>\c@secnumdepth
    \let\@svsec\@empty
  \else
    \refstepcounter{#1}%
    \ifnum #2=2
      \edef\@svsec{\thesubsection.\hskip 0.6em}%
    \else
      \edef\@svsec{\csname the#1\endcsname.\hskip 0.6em}%
    \fi
  \fi
  \@tempskipa #5\relax
  \ifdim \@tempskipa>\z@
    \begingroup #6\relax
      \@hangfrom{\hskip #3\relax\@svsec}%
      \interlinepenalty \@M #8\par
    \endgroup
  \else
    \def\@svsechd{#6\hskip #3\relax
      \@svsec #8}%
  \fi
  \csname #1mark\endcsname{#7}%
  \addcontentsline{toc}{#1}{%
    \ifnum #2>\c@secnumdepth \protect\numberline{}\fi
    #7}%
}

\def\subsection{\@startsection{subsection}{2}{\z@}%
  {1.25ex plus .5ex minus .5ex}% space above subsection
  {1.0ex plus .2ex}% space below subsection
  {\normalfont\normalsize\bfseries}}

\makeatother

\begin{document}

\title{Domain Incremental Learning for Pandemic-Resilient Chest X-Ray Analysis}

\author{
\IEEEauthorblockN{Danu Kim\\[0.3em]}
\IEEEauthorblockA{Korea International School, Jeju Campus, Jeju-do 63644, Republic of Korea\\
Email: dukim27@kis.ac}
}

\maketitle
% ----- compact one-column paragraph under author block -----
\vspace{-5.5em} % tighten vertical space (adjust as needed)
\begin{strip}
\vspace{-3em}   % further reduce top space inside strip
\begin{center}
\textbf{Abstract}
\end{center}

\begin{minipage}{0.95\textwidth}
\justifying
Deep learning models achieved high accuracy in pneumonia detection from chest X-rays. 
However, their generalization across clinical domains remains limited due to variations in imaging devices, 
acquisition protocols, and institutional conditions. 
This study introduces a replay-based domain-incremental continual learning designed to enable continual adaptation to cross-domain variations without catastrophic forgetting. 
The proposed method incorporates a class-aware balanced replay to maintain balanced class representation within a constrained memory and a class-aware loss to dynamically reweight class imbalance during training. 
Experiments conducted on a domain-shifted PneumoniaMNIST dataset consisting of five simulated domains demonstrate that the proposed method achieves an average accuracy of 88.66\%, outperforming Experience Replay, Fine-Tuning, and Joint Training baselines. 
These findings highlight the efficacy of the proposed approach in achieving robust and consistent pneumonia detection across clinical environment variations.
\end{minipage}
\vspace{0.5em} % reduce bottom spacing
\end{strip}
\vspace{0.51em}
% ----- end paragraph -----

\section{Introduction}
\vspace{0.5em}
Deep learning has achieved remarkable success in pneumonia detection from chest X-rays. Early research focused on creating large annotated datasets. Demner-Fushman et al.~\cite{demner2016} released 8,121 images with structured reports, establishing a foundation for standardized radiology corpora. Wang et al.~\cite{wang2017} later introduced ChestX-ray8, containing over 100,000 labeled images across eight thoracic diseases, setting a benchmark for large-scale classification. Building on these datasets, models such as CheXNet~\cite{rajpurkar2017} achieved radiologist-level accuracy, while fine-tuned architectures like VGG16~\cite{aledhari2019} outperformed deeper networks such as ResNet-50. Following these foundational 
studies, numerous subsequent works have been developed to 
further advance chest X-ray analysis.

Despite these advances, most deep learning models assume that training and test data come from the same source, a condition rarely met in clinical practice. Differences in scanners, hospital sites, and patient populations cause cross-domain variations and dataset shifts, degrading model performance. Retraining from scratch can restore accuracy, but privacy and data access restrictions make this impractical. During pandemics, imaging distributions shift rapidly due to new disease patterns and changing protocols. Thus, developing models that remain stable and effective under evolving conditions is critical for pandemic-resilient AI.

Continual learning (CL) enables models to learn new data without forgetting prior knowledge. Van de Ven et al.~\cite{vandeven2022} define three CL scenarios: task-incremental, class-incremental, and domain-incremental learning. The latter is most relevant for medical imaging, where diagnostic tasks remain constant but input data distributions change. Replay-based CL algorithms, such as Experience Replay (ER)~\cite{chaudhry2019}, are effective across these settings.

\begin{figure}[!t]
\centering
\includegraphics[width=\columnwidth]{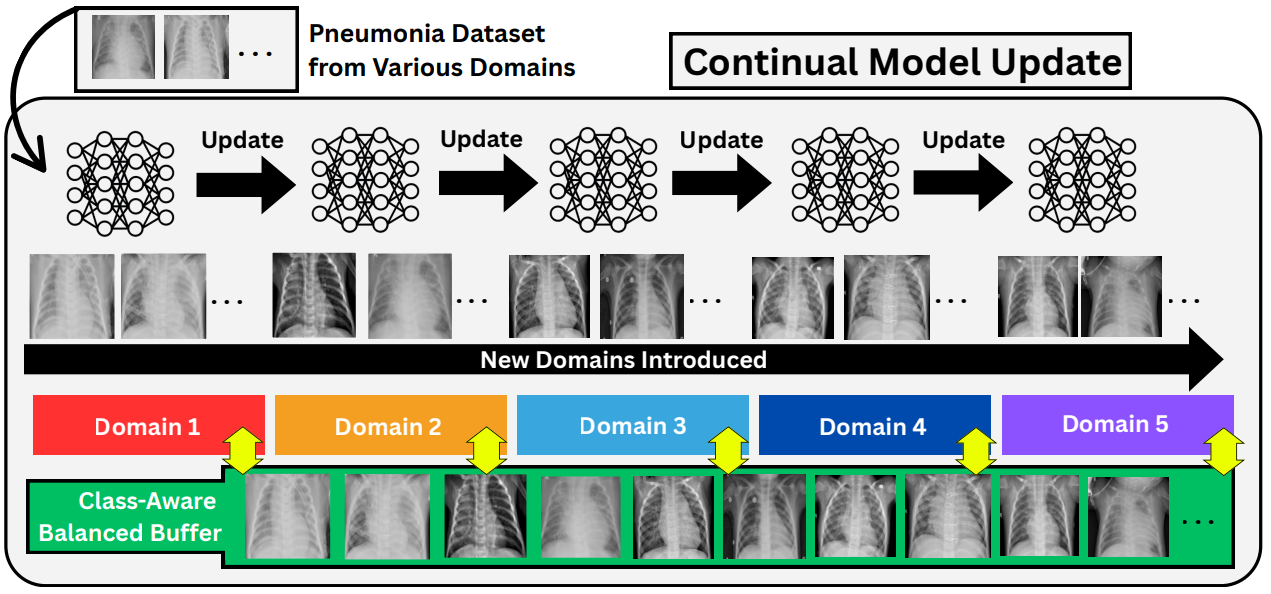}
\caption{System diagram of the proposed method.}
\label{fig:system}
\end{figure}

Within this context, domain-incremental learning is vital for medical imaging, where input distributions shift across devices or clinical environments. This study introduces a domain-incremental learning method integrating class-aware balanced replay and class-aware loss reweighting to mitigate catastrophic forgetting and maintain robust performance across sequential domain changes.

\section{Proposed Method}
\vspace{0.5em}

Fig.~\ref{fig:system} illustrates the system diagram of the proposed method for pneumonia detection using chest X-rays. It depicts a 
replay-based continual learning process where a deep learning 
model is continuously updated as new chest X-ray data from 
different domains (from Domain 1 to Domain 5; e.g., 
hospitals, scanners, or populations) become available. This 
simulates the real-world evolution of data sources over time, 
especially relevant in dynamic medical environments during 
pandemics.

The class-aware balanced buffer stores a small, 
class-specific balanced subset of past samples from each 
domain and class to replay during training. This experience 
replay mechanism ensures the model does not forget previous 
knowledge when learning new domains as imaging conditions 
change.

\subsection{Class-Aware Balanced Replay Algorithm}
The proposed method maintains a replay buffer of past samples. At each step, the model trains on a combined batch containing new data $(x_i, y_i)$ and replayed samples $(x_{replay}, y_{replay})$, selected via class-aware sampling to preserve class balance. The buffer stores a class-specific balanced subset of past samples to ensure older knowledge is retained.

We use a class-aware balanced buffer to keep the memory 
balanced as the model learns new data over time. It combines 
two simple ideas: keeping each class equally represented in the 
buffer and drawing balanced samples during training. This is 
crucial in medical imaging, where class imbalance such as 
fewer pneumonia cases can bias the model toward majority 
classes.

When a new example $(x_i, y_i)$ arrives, it is added to its class buffer $B_{y_i}$ (pneumonia or normal). If that buffer is full, a stored example is randomly replaced:
\[
B_{y_i} \leftarrow
\begin{cases}
\text{append}(x_i, y_i), & \text{if~} |B_{y_i}| < K \\
\text{replace}(B_{y_i}, (x_i, y_i)), & \text{otherwise}
\end{cases}
\]
where $K$ is the per-class capacity. 

During training, the buffer is used to replay past data. 
The sampling function selects an equal number of examples from each class 
to ensure that rare classes are not overlooked. 
If some classes have fewer samples, the remaining batch slots are filled 
by randomly selecting from the entire buffer,
\[
ReplaySet \leftarrow \bigcup_{y_i} RandSample(B_{y_i}, k)
\]
where $k$ is the number of samples drawn from each class 
to form the replay set.

\subsection{Calculating Class-Aware Loss}
The balanced buffer approach keeps the memory compact 
yet evenly balanced across classes, helping the model retain 
prior knowledge while learning new data.

Although the buffer promotes balanced replay, new batches 
may still have uneven class distributions. To prevent overfitting 
to majority classes, we apply class-aware loss, which adjusts 
each class’s contribution to the loss according to its frequency 
in the combined batch of the new incoming and replayed data. 
Classes with fewer samples receive higher weights, ensuring 
balanced learning. The training objective is a class-weighted 
cross-entropy loss:

\[
L = \text{CrossEntropy}(\hat{y}, y, C_{weights})
\]
where $C_{weights}$ are dynamically updated per batch. Classes with fewer samples receive higher weights, maintaining fairness and performance stability.

\section{Experimental Results}
\subsection{Experimental Setup}
\textbf{Datasets.} We evaluate the method on a domain-shifted PneumoniaMNIST~\cite{yang2023} dataset (5,856 pediatric chest X-rays, 28$\times$28 pixels) for binary pneumonia classification. Five domains simulate real-world variations: Base, LowDose, Portable, Anatomical, and Institutional. They are learned sequentially: Base $\rightarrow$ LowDose $\rightarrow$ Portable $\rightarrow$ Anatomical $\rightarrow$ Institutional. Base represents the 
original PneumoniaMNIST, LowDose simulates low radiation 
dose imaging, Portable simulates mobile X-ray scans with blur, 
Anatomical introduces differences in body habitus, and 
Institutional reflects inter-hospital variance in image appearance.

\textbf{Models.} A compact two-stage CNN is used; each stage has a 3$\times$3 convolution, ReLU, and 2$\times$2 max-pooling. The flattened output (3,136 units) connects to a 128-unit ReLU layer and two output nodes. We compare the proposed method (CNN + Class-Aware Replay/Loss) against Experience Replay (ER; CNN + Reservoir Buffer), Fine-Tuning, and Joint Training baselines. All models are trained using Adam (lr=0.001), batch size 32, replay buffer 1,000, and 50 epochs per domain.

\textbf{Metrics.} Evaluation metrics include average accuracy and average forgetting, following standard CL evaluation, with all results reported as the average over three independent runs.

\subsection{Performance Comparison}

\begin{figure}[!t]
\centering
\includegraphics[width=\columnwidth]{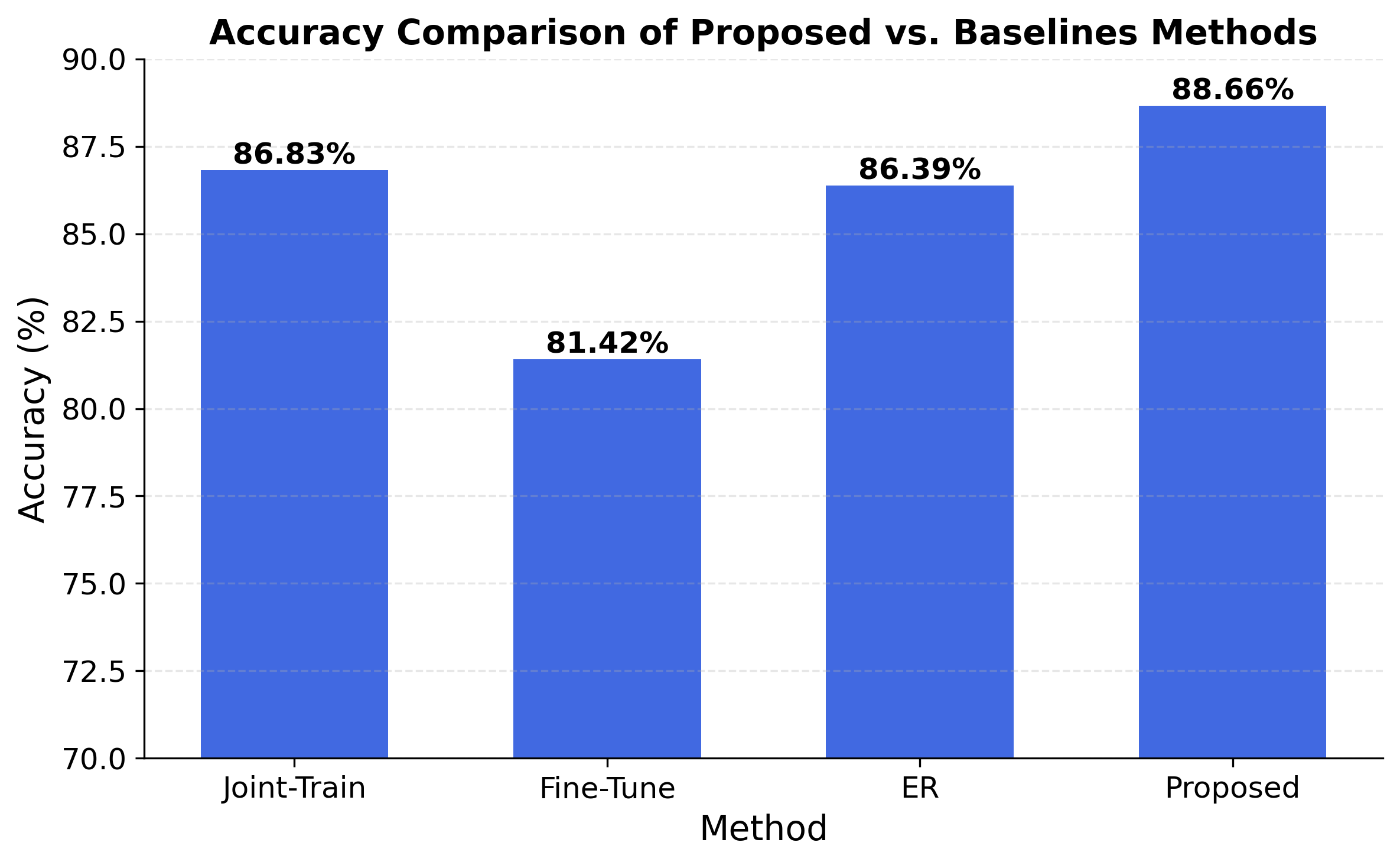}
\caption{Performance comparison of the proposed method with baseline methods.}
\label{fig:accuracy}
\end{figure}

Fig.~\ref{fig:accuracy} shows that the proposed method achieves the 
highest average accuracy of 88.66\%, surpassing ER (86.39\%), 
Joint Training (86.83\%), and Fine-Tuning (81.42\%). Although 
the accuracy differences are modest, the overall improvement 
demonstrates the robustness and adaptability of our proposed method across domains. Compared to the baselines, the method consistently 
maintains higher accuracy across all domains, confirming its 
effectiveness in enabling stable domain-incremental learning.

\begin{table}[!t]
\centering
\caption{Accuracy changes of the proposed method as new domains introduced and learned.}
\begin{tabular}{lr}
\toprule
\textbf{Domain} & \textbf{Accuracy Trends (\%)} \\
\midrule
Base & 85.10 $\rightarrow$ 86.86 $\rightarrow$ 86.81 $\rightarrow$ 92.10 $\rightarrow$ 89.96 \\
LowDose & 84.78 $\rightarrow$ 84.51 $\rightarrow$ 85.31 $\rightarrow$ 85.42 \\
Portable & 85.95 $\rightarrow$ 90.49 $\rightarrow$ 90.81 \\
Anatomical & 89.26 $\rightarrow$ 87.45 \\
Institutional & 89.69 \\
\bottomrule
\end{tabular}
\label{tab:domains}
\end{table}

Table~\ref{tab:domains} illustrates that across five sequential domains, the 
proposed method exhibits stable continual learning. Accuracy in 
the Base domain increases from 85.10\% to 92.10\% before 
slightly decreasing to 89.96\%, indicating strong retention with 
minimal decline. LowDose remains steady (84.78$\rightarrow$85.42\%), 
while Portable improves notably (85.95$\rightarrow$90.81\%), reflecting 
consistent learning. Anatomical shows a slight decrease (89.26$\rightarrow$87.45\%), suggesting minor forgetting as new domains are 
introduced, and Institutional reaches 89.69\% as the final 
domain. Overall, the results demonstrate stable retention and 
controlled forgetting, validating the robustness of the proposed 
method across sequential domains.

\begin{figure}[!t]
\centering
\includegraphics[width=\columnwidth]{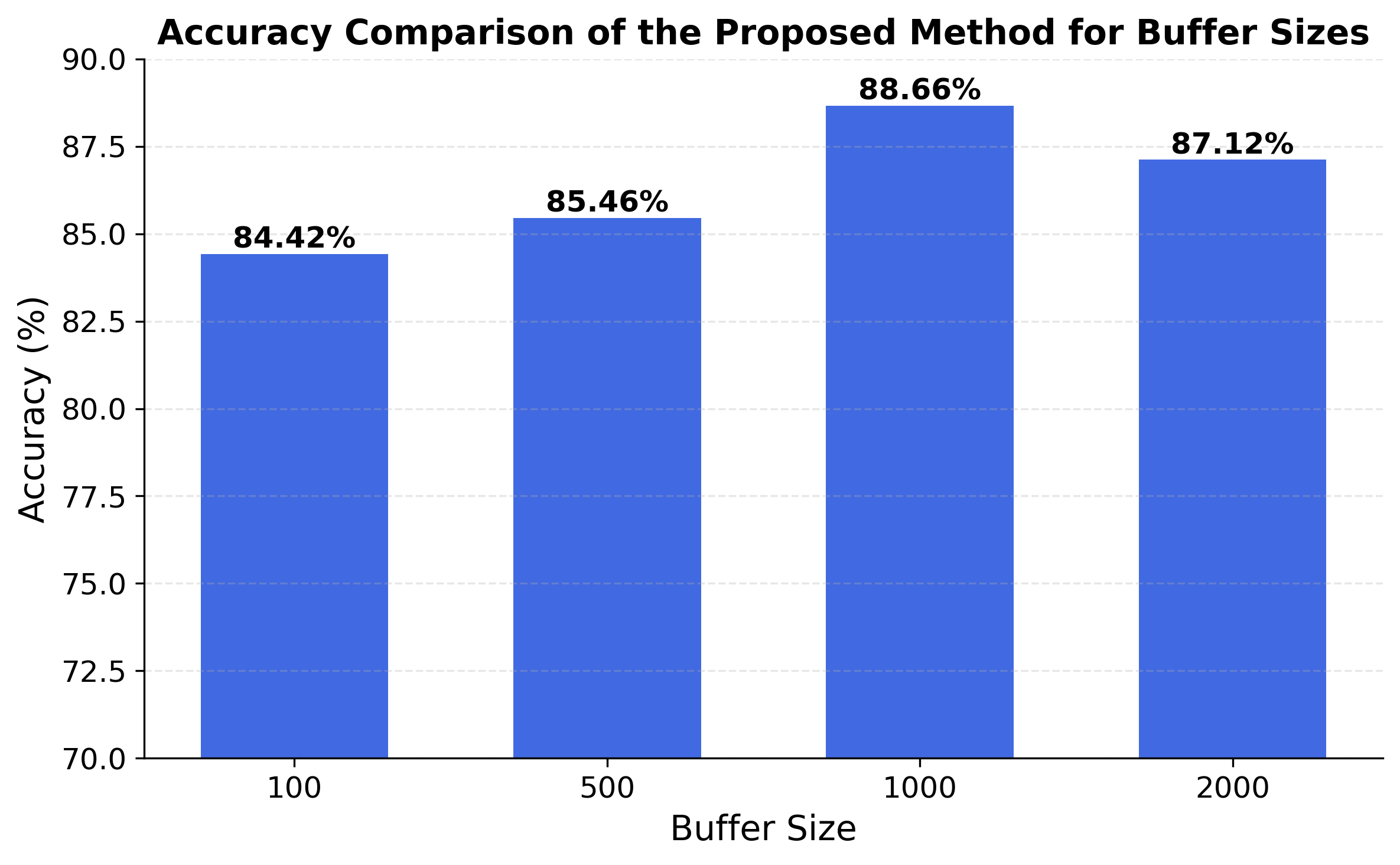}
\caption{Accuracy of the proposed method for buffer sizes.}
\label{fig:buffer}
\end{figure}

Fig.~\ref{fig:buffer} shows that accuracy increases steadily from 84.42\% at a buffer size of 100 to a peak of 88.66\% at 1,000, 
followed by a slight decrease to 87.12\% at 2,000. This trend 
indicates that expanding the buffer generally improves 
performance, but the benefit diminishes as redundancy grows. 
The proposed method maintains per-class balance within the 
buffer, and ensures uniform exposure during training. Together, 
these components enhance data diversity and fairness, enabling 
efficient use of limited memory. The observed peak accuracy 
at a buffer size of 1,000 suggests an effective trade-off 
between replay quality and buffer capacity.

We analyze how different optimization strategies affect continual learning stability. As shown in Table~\ref{tab:optim}, Adam outperforms SGD in accuracy 
(88.66\% vs. 85.37\%), reflecting its adaptive learning advantage 
in continual learning. However, SGD exhibits slightly lower 
forgetting (0.74\% vs. 0.79\% for Adam), indicating stronger 
knowledge retention. SGD also demonstrates greater stability, 
with smaller standard deviations in accuracy (±0.53 vs. ±0.98) 
and forgetting (±1.08 vs. ±0.33). Overall, Adam achieves 
higher peak performance, while SGD provides more consistent 
results.

\begin{table}[!t]
\centering
\caption{Comparison for two optimizers: Adam vs. SGD}
\begin{tabular}{lcc}
\toprule
\textbf{Optimizer} & \textbf{Accuracy (\%)} & \textbf{Forgetting (\%)} \\
\midrule
Adam & 88.66 ($\pm$0.98) & 0.79 ($\pm$1.08) \\
SGD & 85.37 ($\pm$0.53) & 0.74 ($\pm$0.33) \\
\bottomrule
\end{tabular}
\label{tab:optim}
\end{table}

\begin{table}[!t]
\centering
\caption{Comparison based on training models}
\begin{tabular}{lcc}
\toprule
\textbf{Model} & \textbf{Accuracy (\%)} & \textbf{Forgetting (\%)} \\
\midrule
CNN & 88.66 ($\pm$0.98) & 0.79 ($\pm$1.08) \\
MobileNetV2 & 86.53 ($\pm$0.84) & 1.74 ($\pm$0.64) \\
\bottomrule
\end{tabular}
\label{tab:model}
\end{table}

Finally, we examine how model complexity influences continual learning stability. Comparing the CNN and MobileNetV2 reveals whether simpler architectures better retain prior knowledge or whether deeper ones provide stronger generalization under domain variations. As shown in Table~\ref{tab:model}, the CNN achieves higher accuracy (88.66\%) and lower forgetting (0.79\%) compared to MobileNetV2 (86.53\%, 1.74\%), indicating stronger performance and greater resistance to catastrophic forgetting. The higher forgetting observed in MobileNetV2 may result from its deeper and more complex architecture, which can make it less stable when adapting to new domain tasks. In contrast, the lighter CNN demonstrates not only efficient learning but also better retention of prior knowledge. The slightly higher variability in the CNN’s forgetting suggests greater adaptability and flexibility in continual learning.

\section{Conclusion}
\vspace{0.5em}
In this work, we propose a replay-based domain-incremental 
learning algorithm for classifying pneumonia from 
chest X-rays, addressing the challenge of adapting to domain 
changes in real-world clinical imaging environments. 
The proposed method achieves high accuracy, low forgetting, 
and robust performance under varying imaging conditions 
through the use of a class-aware balanced replay and 
class-aware loss.

\bibliographystyle{IEEEtran}

\end{document}